\documentclass[11pt,a4paper]{article}
\usepackage[hyperref]{acl2020}
\usepackage{times}
\usepackage{latexsym}
\usepackage{url}
\usepackage{amsmath}
\usepackage[linesnumbered,lined,ruled,commentsnumbered]{algorithm2e}
\usepackage{adjustbox}
\usepackage{natbib}
\usepackage{CJKutf8}
\usepackage{multirow}

\usepackage{xcolor}

\aclfinalcopy 
\title{A Reinforced Generation of Adversarial Examples for Neural Machine Translation}

\author{Wei Zou$^{1}$\quad Shujian Huang$^{1}$\quad Jun Xie$^{2}$\quad Xinyu Dai$^{1}$\quad Jiajun Chen$^{1}$ \\
$^{1}$National Key Laboratory for Novel Software Technology, Nanjing University, China\\
$^{2}$Tencent Technology Co, China \\
\texttt{zouw@smail.nju.edu.cn},\ \texttt{\{huangsj,daixinyu,chenjj\}@nju.edu.cn}\ \\
\texttt{stiffxie@tencent.com} \\
}

\date{}

\begin{document}
\maketitle

\begin{abstract}
Neural machine translation systems tend to fail on less decent inputs despite its significant efficacy, which may significantly harm the credibility of these systems—fathoming how and when neural-based systems fail in such cases is critical for industrial maintenance. Instead of collecting and analyzing bad cases using limited handcrafted error features, here we investigate this issue by generating adversarial examples via a new paradigm based on reinforcement learning. Our paradigm could expose pitfalls for a given performance metric, e.g., BLEU, and could target any given neural machine translation architecture. We conduct experiments of adversarial attacks on two mainstream neural machine translation architectures, RNN-search, and Transformer. The results show that our method efficiently produces stable attacks with meaning-preserving adversarial examples. We also present a qualitative and quantitative analysis for the preference pattern of the attack, demonstrating its capability of pitfall exposure.
\end{abstract}

\section{Introduction}  \label{introduction}


\noindent Neural machine translation (NMT) based on the encoder-decoder framework, such as RNN-Search \citep[RNNSearch]{Bahdanau2015Neural, Luong2015Effective}
or Transformer \citep[Transformer]{vaswani2017attention}, has achieved remarkable progress and become a de-facto in various machine translation applications. However, there are still pitfalls for a well-trained neural translation system, especially when applied to less decent real-world inputs compared to training data \citep{belinkov2017synthetic}. For example, typos may severely deteriorate system outputs (Table \ref{tab:brittle_nmt}). 
Moreover, recent studies show that a neural machine translation system can also be broken by noisy synthetic inputs \citep{belinkov2017synthetic,lee2018hallucinations}.
Due to the black-box nature of a neural system, it has been a challenge to fathom when and how the system tends to fail.

\begin{table}[!t]
    \centering
    \resizebox{0.45\textwidth}{!}{
        \begin{tabular}{c|l}
        \hline
        \textbf{in} & \begin{CJK}{UTF8}{gkai}耶路撒冷发生自杀\textit{\textbf{爆炸}} 事件 \end{CJK}  \\
        \hline
        \textbf{out} & suicide bombing in jerusalem \\
        \hline
        \textbf{in} & \begin{CJK}{UTF8}{gkai}耶路撒冷发生自杀\textit{\textbf{爆}}事件\end{CJK} \\
        \hline
        \textbf{out} & \textit{eastern} jerusalem \textit{explores a case of eastern europe} \\
        \hline
        \end{tabular}
    }
    \caption{Fragility of neural machine translation. A typo leaving out a Chinese character ``\begin{CJK}{UTF8}{gkai}炸\end{CJK}'' leads to significant errors (noted by italics) in English translation. Both ``\begin{CJK}{UTF8}{gkai}爆\end{CJK}'' and ``\begin{CJK}{UTF8}{gkai}爆炸\end{CJK}'' mean ``bombing'' in English.}
    \label{tab:brittle_nmt}
\end{table}

Intuitively, researchers seek to apprehend such failures by the analysis of handcrafted error indicating features~\citep{zhao2018addressing, karpukhin2019training}. This strategy is costly because it requires expert knowledge for both linguistics and the target neural architecture. Such features are also less applicable because some common errors in deep learning systems are hard to formulate, or very specific to certain architectures.

Instead of designing error features, recent researchers adopt ideas from adversarial learning~\citep{goodfellow2014explaining} to generate adversarial examples 
for mining pitfalls of NLP systems ~\citep{cheng2018seq2sick,ebrahimi2018adversarial,zhao2017generating}.
Adversarial examples are minor perturbed inputs that keep the semantic meaning, yet yield degraded outputs.
The generation of valid adversarial examples provides tools for error analysis that is interpretable for ordinary users, which can contribute to system maintenance.
Though it has achieved success concerning continuous input, e.g., images, there are following major issues for NLP tasks. 

First, it is non-trivial to generate valid discrete tokens for natural language, e.g., words or characters. 
\newcite{cheng2018seq2sick} follow \newcite{goodfellow2014explaining} to learn noised representation then sample tokens accordingly. However, there is no guaranteed correspondence between arbitrary representation and valid tokens. Therefore, it may generate tokens departing from learned representation, which undermines the generation.
\newcite{ebrahimi2018adversarial} turns to a search paradigm by a brute-force search for direct perturbations on the token level.
To lead the search, a gradient-based \textit{surrogate} loss must be designed upon every token modification by given target annotations. 
However, this paradigm is inefficient due to the formidable computation for gradients. 
Furthermore, surrogate losses defined upon each token by targets requires high-quality targets, and risks being invalidated by any perturbation that changes tokenization.

Another issue is to keep the semantics of original inputs. 
Different from the fact that minor noises on images do not change the semantics, sampling discrete tokens from arbitrary perturbed representation \citep{cheng2018seq2sick} may generate tokens with different semantics and lead to ill-perturbed samples (Table \ref{tab:unwanted_perturb}).
Searching for the perturbed input also requires a semantic constraint of the search space, for which handcrafted constraints are employed \citep{ebrahimi2018adversarial}.
Though constraints can also be introduced by multitask modeling with additional annotations \citep{zhao2017generating}, this is still not sufficient for tasks requiring strict semantic equivalence, such as machine translation. 

\begin{table}[!t]
    \centering
    \resizebox{0.4\textwidth}{!}{
        \begin{tabular}{c|l}
        \hline
        \textbf{in} & Two man are playing on the street corner. \\
        \hline
        \textbf{perturbed in} & Two man are playing \textit{\textbf{frisbee in the park}}. \\
        \hline
        \textbf{out} & Zwei Männer spielen an einer Straßenecke. \\
        \hline
        \textbf{perturbed out} & Zwei Männer spielen frisbee im park. \\
        \hline
        \end{tabular}
    }
    \caption{Example of undesirable perturbation in adversarial examples for machine translation in  \citep{zhao2017generating}, though it yields very different output compare to the origin, it does not indicate system malfunction.}
    \label{tab:unwanted_perturb}
\end{table}

In this paper, we adopt a novel paradigm that generates more reasonable tokens and secures semantic constraints as much as possible. We summarize our contributions as the following:
\begin{itemize}

\item 
We introduce a reinforcement learning \citep[RL]{sutton2018reinforcement} paradigm with a discriminator as the terminal signal in its environment to further constrain semantics. This paradigm learns to apply discrete perturbations on the token level, aiming for direct translation metric degradation. 
Experiments show that our approach not only achieves semantically constrained adversarial examples but also leads to effective attacks for machine translation.

\item
Our paradigm can achieve the adversarial example generation with outclassed efficiency by only given source data.
Since our method is model-agnostic and free of handcrafted error feature targeting architectures, it is also viable among different machine translation models.

\item 
We also present some analysis upon the state-of-the-art Transformer based on its attack, showing our method's competence in system pitfall exposure. 
\end{itemize}

\section{Preliminaries}

\begin{figure*}[t]
    \centering
    \includegraphics[width=1.0\textwidth]{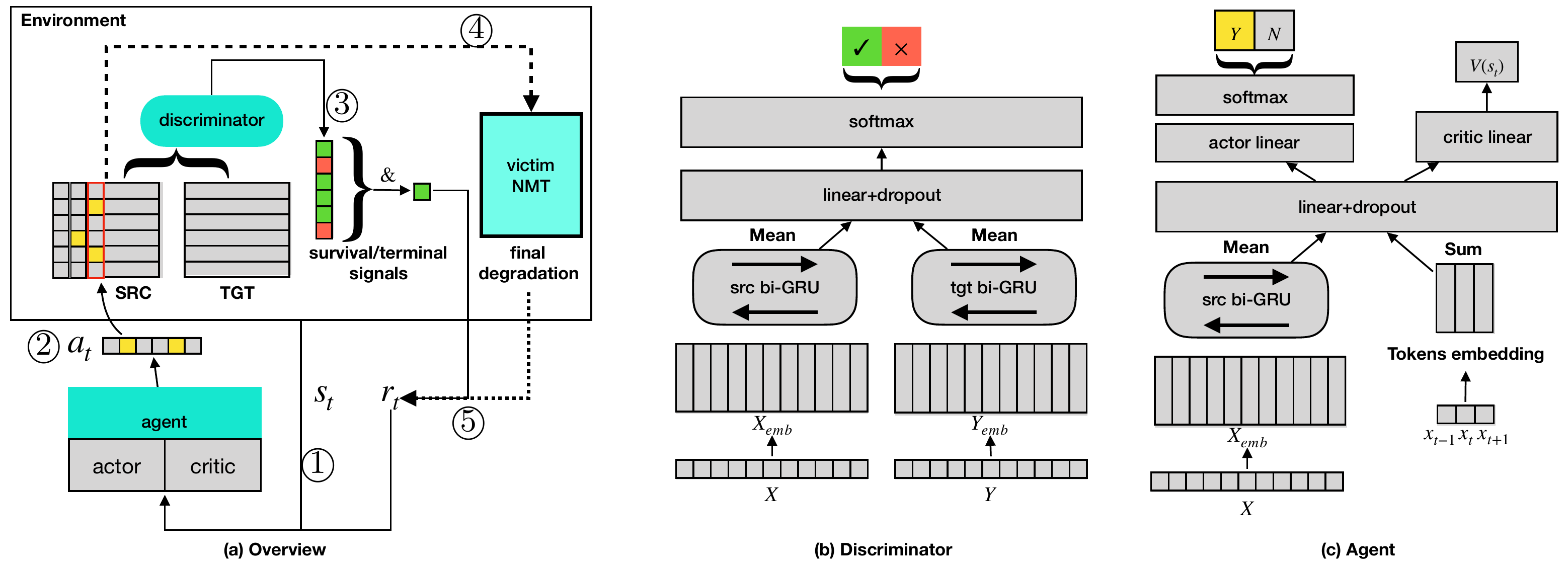}
    \caption{[a] Overview of our RL architecture. \textcircled{1} Environment states are processed as inputs for agent; \textcircled{2} agent yield modification upon $SRC$ in environment; \textcircled{3} determine survival and step reward of environment; \textcircled{4} determine degradation with victim NMT as episodic reward; \textcircled{5} update agent with total rewards. During a training episode, we loop \textcircled{1} to \textcircled{3} and accumulate step rewards until environment terminates. Dash line indicates execution at the end of an episode. [b] Architecture of discriminator. [c] Architecture of agent. } 
    \label{fig:model_overview}
\end{figure*}

\subsection{Neural Machine Translation}
The most popular architectures for neural machine translation are RNN-search \citep{Bahdanau2015Neural} and Transformer \citep{vaswani2017attention}.
They share the paradigm to learn the conditional probability $P(Y|X)$ of a target translation $Y=[y_1,y_2,...,y_m]$ given a source input $X=[x_1,x_2, ..., x_n]$. A typical NMT architecture consists of an encoder, a decoder and attention networks. The encoder encodes the source embedding $X_{emb}=[emb_1,emb_2,...emb_n]$ into hidden representation $H=[h_1,h_2,...,h_n]$.
Then a decoder $f_{dec}$ with attention network attentively accesses $H$ for an auto-regressive generation of each $y_i$ until the end of sequence symbol (EOS) is generated:
\begin{align}
    P(y_i|y_{<i},X)&=\text{softmax}(f_{dec}(y_{i-1}, s_t,c_t; \theta_{dec})) \label{eq:nmt_output}
\end{align}
where $c_t$ is the attentive result for current decoder state $s_t$ given $H$.

\subsection{Actor-Critic for Reinforcement Learning}
Reinforcement learning \citep[RL]{sutton2018reinforcement} is a widely used machine learning technique following the paradigm of \textit{explore and exploit}, which is apt for unsupervised policy learning in many challenging tasks (e.g., games \citep{mnih2015human}). It is also used for direct optimization for non-differentiable learning objectives \citep{wu2018study,bahdanau2016actor} in NLP. 

Actor-critic \citep{konda2000actor} is one of the most popular RL architectures where the agent consists of a separate policy and value networks called actor and critic. 
They both take in environment state $s_t$ at each time step as input, while actor determines an action $a_t$ among possible action set $\mathcal{A}$ and critic yields value estimation $V_t(s_t)$ . 
In general, the agent is trained to maximize discounted rewards $R_t=\sum_{i=0}^\infty{\gamma^i r_{t+i}}$ for each state, where $\gamma\in(0,1]$ is the discount factor. Such goal can be further derived as individual losses applied to actor and critic.
Thus the actor policy loss $L^{\pi}$ on step $t$ is: 
\begin{align}
    L^{\pi}_t(\theta_{\pi})=\log P(a_t|s_t)A_t(s_t, a_t); a_t\in \mathcal{A}  \label{eq:actor_loss}
\end{align}
where $\theta_{\pi}$ denotes actor parameters, $A_t(s_t, a_t)$ denotes general advantage function \citep{schulman2015high} on state $s_t$ for action $a_t$ given by $\sum^{k-1}_{i=0} \gamma^ir_{t+i}+\gamma^kV(s_{t+k})-V(s_t) $, which can be further derived as:
\begin{align}
    A_t(s_t,a_t) = \gamma A_{t+1}(s_{t+1}, a_{t+1})+ r_t \nonumber \\ 
    +\gamma V_{t+1}(s_{t+1})-V_t(s_t)
    \label{eq:gae}
\end{align}
Meanwhile, critic learns to estimate $R_t$ via minimizing a temporal difference loss $L^{v}$ on each step $t$:
\begin{align}
    L^{v}_t(\theta_v)=\frac{1}{2}(r_t+\gamma R_{t+1}- V_t(s_t))^2 \label{eq:critic_loss}
\end{align}
where $\theta_v$ denotes critic parameter. 

Usually, the training is regularized by maximizing policy entropy  $H^{\pi}$  to avoid exploration failure before exploiting optimum policy \citep{ziebart2010modeling}. Thus the total loss becomes:
\begin{align}
    L(\theta)=\sum_t(\alpha L^v_t - L^{\pi}_t - \beta H^{\pi}(\cdot|s_t))
    \label{eq:overall_loss}
\end{align}
where $\alpha$ and $\beta$ are hyperparameters for value loss and entropy coefficients. 

\subsection{adversarial examples in NLP}
A general adversarial example generation can be described as the learning process to find a perturbation $\delta$ on input $X$ that maximize system degradation $L_{adv}$ within a certain constraint $C(\delta)$:
\begin{align}
    \underset{\delta}{\mathrm{argmax}} L_{adv}(X+\delta)-\lambda C(\delta)  \label{eq:adversarial_loss_continuous}
\end{align}
where $\lambda$ denotes the constraint coefficient, $ L_{adv}$ is determined by the goal of the attack. 
However, currently effective adversarial generation for NLP is to search by maximizing a surrogate gradient-based loss:
\begin{align}
    \underset{1 \leq i \leq n, x'\in vocab}{\mathrm{argmax}} L_{adv}(x_0, x_1,...x_i'...x_n)
    \label{eq:adversarial_loss_discrete}
\end{align}
where $L_{adv}$ is a differentiable function indicating the adversarial object. Due to its formidable search space, this paradigm simply perturbs on a small ratio of token positions and greedy search by brute force among candidates. 
Note that adversarial example generation is fundamentally different from noised hidden representation in adversarial training \citep{cheng2019robust, sano2019effective}, which is not to be concerned in this work.

\section{Approach}

In this section, we will describe our reinforced learning and generation of adversarial examples (Figure \ref{fig:model_overview}) in detail. 
Overall, the victim model is a part of the \textbf{environment} (denoted as $Env$), which yields rewards indicating overall degradation based on modified inputs.
A reinforced \textbf{agent} learns to modify every source position from left to right sequentially. 
Meanwhile, a \textbf{discriminator} in $Env$ provides every-step survival signals by determining whether SRC is ill-perturbed.

\subsection{Environment}
We encapsulate the victim translation model with a discriminative reward process as an $Env$ for a reinforced agent to interact.

\subsubsection{Environment State}
The state of the $Env$ is described as $s_t = (SRC, t)$, where $SRC=[src_0, src1,...,src_N]$ are $N$ sequences processed by victim model's vocabulary and tokenization. 
Each sequence $src_i=[x_1, x_2, ..., x_n]$ is concatenated with $BOS, EOS$, which indicate the begin and end of the sequence, then padded to same length. 
Time step $t \in[1,n]$ also indicates the token position to be perturbed by the agent. 
$Env$ will consecutively loop for all token positions and update $s_t$ based on the agent's modification.
$Env$ also yields reward signals until the end or intermediately terminated. 
That is, \textit{all} sequences in $SRC$ are determined by $D$ as ill-perturbed during the reward process. 
Once the $Env$ terminates, it finishes the current episode and reset its state with a new batch of sequences as $SRC$.

\subsubsection{Reward Process with Discriminator}  \label{subsec:Reward_process}
The reward process is \textit{only} used during training. It consists of a survival reward $r_{\text{s}}$ on every step and a final degradation $r_{\text{d}}$ concerning an overall metric if the agent survives till the end. Overall, we have:
\begin{align}
\centering
r_t=
\begin{cases}
-1, \; \text{terminated} \\
\frac{1}{N}{\sum_N a \cdot r_{\text{s}}}, \;\text{survive}\And t\in [1,n) \\
\frac{1}{N}{\sum_N (a \cdot r_{\text{s}}+ b \cdot r_{\text{d}})}, \;\text{survive}\And t=n 
\label{eq:rewards}
\end{cases}
\end{align}
where $a,b$ are hyper parameters that keeps the overall $r_{\text{s}}$ and $r_{\text{d}}$ within similar magnitude.

Instead of direct optimization of the constrained adversarial loss in Eq.\ref{eq:adversarial_loss_continuous}, we model discriminator $D$'s output as survival rewards similar to that in gaming \citep{mnih2015human}.
That is, the agent must survive for its goal by also fooling $D$, which attempts to terminate ill-perturbed modifications.
We define an ill-perturbed source by determining whether it still matches the original target $tgt$. 

\paragraph{Discriminator} As it is shown in Figure \ref{fig:model_overview}(b), discriminator $D$ consists of bi-directional GRU encoders for both source and target sequence. Their corresponding representation is averaged and concatenated before passed to a feedforward layer with dropout. Finally, the output distribution is calculated by a softmax layer.
Once $D$ determines the pair as positive, its corresponding possibility is regarded as the reward, otherwise 0:
\begin{align}
\centering
r_{\text{s}}=
\begin{cases}
P(\text{positive}|(src',tgt); \theta_d), & \text{positive} \\
0, & \text{otherwise}\\
\end{cases}
\label{eq:r_degrade}
\end{align}

As long as the environment survives, it yields averaged reward among samples from $SRC$ (Eq.\ref{eq:rewards}) to mitigate rewards' fluctuation that destabilize training. 

\paragraph{Discriminator Training}
Similar to GAN training, the environment's $D$ must update as the agent updates. During its training, the agent's parameter is freezed to provide training samples. 
For every $D$'s training epoch, we randomly choose half of the batch and perturb its source using the current agent as negative samples. 
During $D$'s updates, we randomly generate a new batch of pairs from parallel data likewise to test its accuracy. 
$D$ is updated at most $step_D$ epochs, or until its test accuracy reaches ${acc\_bound}$.

$Env$ \textit{only}\footnote{It is commonly accepted that frequent negative rewards result in agents' tendency to regard zero-reward as optimum and fail exploration, which further leads to training failure.} yields -1
as overall terminal rewards when all sequences in $SRC$ are \textit{intermediately} terminated. For samples classified as negative during survival, their follow-up rewards and actions are masked as 0. 
If the agent survives until the end, $Env$ yields additional averaged $r_{\text{d}}$ as final rewards for an episode. 
We follow \newcite{michel2019evaluation} to adopt relative degradation:
\begin{align}
\centering
    r_{\text{d}}=\frac{\text{score}(y,refs)-\text{score}(y',refs)}{\text{score}(y,refs)}
    \label{eq:RD}
\end{align}
where $y$ and $y'$ denote original and perturbed output, $refs$ are references, and $score$ is a translation metric. If $score(y,refs)$ is zero, we return zero as $ r_{\text{d}}$. 
To calculate $\text{score}$ we \textit{retokenize} perturbed $SRC$ by victim models vocabulary and tokenizer before translation.

\subsection{Agent}
As it is shown in Figure \ref{fig:model_overview} (c), the agent's actor and critic share the same input layers and encoder, but later processed by individual feedforward layers and output layers. Actor takes in $SRC$ and current token with its surrounding $(x_{t-1}, x_t, x_{t+1})$, then yields a binary distribution to determine whether to attack a token on step $t$, while critic emits a value $V(s_t)$ for every state.
Once the actor decides to perturb a specific token, this token will be replaced by another token in its candidate set.

\paragraph{Candidate Set}
We collect at most $K$ candidates for each token in the victim's vocabulary within a distance of $\epsilon$. $\epsilon$ is the averaged Euclidean distance of $K$-nearest embedding for all tokens in victim vocabulary. 
We note that there shall always be candidates for a token in test scenarios that are beyond victim's vocabulary, for those without a nearby candidate, we assign UNK as its candidate.
Once the agent chooses to replace a token with UNK, we follow \newcite{michel2019evaluation} to present a valid token that is also UNK to the victim's vocabulary.

\paragraph{Agent Training}
The agent is trained by algorithm in appendix A.
Since the agent is required to explore with stochastic policy during training, it will first sample based on its actor's output distribution on whether to perturb the current position, then randomly choose among its candidates.
The agent and discriminator take turns to update. 
We assume the training is converged when test accuracy for $D$ does not reach over a certain value within certain continuous learning rounds of agent and discriminator.
\paragraph{Agent Inference}
To generate adversarial examples, the agent will take in source sequences and perturb on each position based on the actor's output from left to right, then choose the nearest candidate. 
As the agent's critic learns to estimate expected future rewards for a step, only when it yields positive value will agent perturb, otherwise it indicates an undesirable perturbation; thus, the agent is muted.


\section{Experiments}

\subsection{Data Sets}
We test our adversarial example generations on Zh$\rightarrow$En, En$\rightarrow$Fr, and En$\rightarrow$De translation tasks, which provide relatively strong baselines for victim models and mass test samples. 

We train our agent using only parallel data that is used for victims' training.
we train on LDC Zh$\rightarrow$En\footnote{ldc2002E18, ldc2003E14, ldc2004T08, ldc2005T06}(1.3M pairs), WMT14 En$\rightarrow$De\footnote{https://nlp.stanford.edu/projects/nmt/} (4.5M pairs) and WMT15 En$\rightarrow$Fr\footnote{Europarl-v7, news-commentary-v10}(2.2M pairs) for victim models respectively.
For subword level translation, we apply byte pair encoding \citep[BPE]{sennrich2015neural} for both source and target languages with the vocabulary size of 37k. We also use join-BPE for En-De and En-Fr experiments with 34k and 33k vocabulary size, respectively.
For word-level translation, we use NLPIR-ICTCLAS and Moses tokenizer for Chinese and English tokenization, respectively.
We adopt 30k as vocabulary size for both source and target language.
We adopt NIST test sets \footnote{MT02,03,04,05,06} for Zh$\rightarrow$En and WMT test sets for En$\rightarrow$De and En$\rightarrow$Fr, then generate adversarial examples for these sources for analysis. 

\subsection{Victim Models}
We choose the state-of-the-art RNN-search and Transformer as victim translation models. 
For RNN-search, we train subword level models and strictly follow the architecture in \newcite{Bahdanau2015Neural}.
As for Transformer, we train both word-level and subword-level model for Zh$\rightarrow$En and only subword-level models for En$\rightarrow$De and En$\rightarrow$Fr with the architecture and the base parameter settings by \newcite{vaswani2017attention}.
For the above models, we apply the same batch scheme and Adam optimizer following \newcite{vaswani2017attention}. We choose MT03, newsdiscuss2015 and newstest2013 for Zh$\rightarrow$En, En$\rightarrow$Fr, En$\rightarrow$De as validation set respectively.

\subsection{Metrics}
We first report attack results both in terms of char-level BLEU (\textit{chrBLEU}) of perturbed source by the origin to indicate modification rate, and relative decrease in target BLEU (\textit{RD}):
\begin{align}
    RD=\frac{\text{BLEU}(y,refs)-\text{BLEU}(y',refs)}{(1-\text{chrBLEU}(x',x))\times \text{BLEU}(y,refs)}
\end{align}
We adopt sacreBLEU \citep{post2018call} to test case-insensitive BLEU on detokenized targets.

As \newcite{michel2019evaluation} suggest, there is a trade-off between achieving high \textit{RD} and maintaining semantic. 
One can achieve rather high \textit{RD} by testing with mismatched references, making degradation less meaningful.
Therefore, we also test source semantic similarity with human evaluation (\textit{HE}) ranging from $0$ to $5$ used by \newcite{michel2019evaluation} by randomly sampling $10\%$ of total sequences mixed with baselines for a double-blind test. 

\subsection{Results}
We implement state-of-the-art adversarial example generation by gradient search \cite{michel2019evaluation} (GS) as a baseline, which can be currently applied to various translation models. 
We also implemented random synthetic noise injection \cite{karpukhin2019training} (RSNI) as an unconstrained contrast. 
Both baselines are required to provide a ratio for the amount of tokens to perturb during an attack, where we present the best results.
Unlike our paradigm can generate on monolingual data, GS also requires target annotations, where we use one of the references to provide a strong baseline.
Note that RSNI can significantly break semantics with distinctly lower \textit{HE} to achieve rather high \textit{RD}, which we do not consider as legit adversarial example generation and noted with ``*'' for exclusion.


\begin{table}[!t]
    \centering
\resizebox{0.45\textwidth}{!}{
\begin{tabular}{llccc}
\hline
                                      & \multicolumn{4}{c}{Zh-En MT02-06}                                                                \\ \cline{2-5} 
                                      & \multicolumn{1}{c}{BLEU} & \multicolumn{1}{c|}{chrBLEU} & \multicolumn{1}{c}{RD$\uparrow$} & HE$\uparrow$            \\ \hline
\multicolumn{1}{l|}{Transformer-word} & 41.16                    & \multicolumn{1}{c|}{-}        &            -            &          -     \\
\multicolumn{1}{l|}{RSNI (0.2)$^*$}    & 29.68                    & \multicolumn{1}{c|}{0.892}   & 2.580$^*$                  & 1.39$^*$          \\
\multicolumn{1}{l|}{RSNI (0.3)$^*$}    & 19.94                    & \multicolumn{1}{c|}{0.781}   & 2.350$^*$                  & 1.10$^*$          \\
\multicolumn{1}{l|}{GS (0.2)}     & 33.46                    & \multicolumn{1}{c|}{0.749}   & 0.746                  & 3.23          \\
\multicolumn{1}{l|}{GS (0.3)}     & 29.86                    & \multicolumn{1}{c|}{0.676}   & 0.847                  & 2.49          \\
\multicolumn{1}{l|}{Ours}             & 33.72                    & \multicolumn{1}{c|}{0.804}   & \textbf{0.952}         & \textbf{3.73} \\ \hline
\multicolumn{1}{l|}{Transformer-BPE}  & 44.06                    & \multicolumn{1}{c|}{-}        &      -                 &        -       \\
\multicolumn{1}{l|}{RSNI (0.2)$^*$}    & 34.44                    & \multicolumn{1}{c|}{0.892}   & 2.019$^*$                  & 1.45$^*$          \\
\multicolumn{1}{l|}{RSNI (0.4)$^*$}    & 25.78                    & \multicolumn{1}{c|}{0.781}   & 1.891$^*$                  & 1.08$^*$          \\
\multicolumn{1}{l|}{GS (0.2)}     & 35.52                    & \multicolumn{1}{c|}{0.823}   & \textbf{1.094}         & \textbf{3.88} \\
\multicolumn{1}{l|}{GS (0.4)}     & 28.18                    & \multicolumn{1}{c|}{0.675}   & 1.004                  & 2.90          \\
\multicolumn{1}{l|}{Ours}             & 35.48                    & \multicolumn{1}{c|}{0.807}   & 1.009                  & 3.79          \\ \hline
\multicolumn{1}{l|}{RNN-search-BPE}   & 40.90                    & \multicolumn{1}{c|}{-}        &  -                      &     -          \\
\multicolumn{1}{l|}{RSNI (0.2)$^*$}    & 32.54                    & \multicolumn{1}{c|}{0.892}   & 1.891$^*$                  & 1.44$^*$              \\
\multicolumn{1}{l|}{RSNI (0.4)$^*$}    & 25.54                    & \multicolumn{1}{c|}{0.781}   & 1.712$^*$                  & 1.36$^*$              \\
\multicolumn{1}{l|}{GS (0.2)}     & 32.94                    & \multicolumn{1}{c|}{0.823}   & \textbf{1.102}         & 3.79          \\
\multicolumn{1}{l|}{GS (0.4)}     & 27.02                    & \multicolumn{1}{c|}{0.678}   & 1.053                  & 2.88          \\
\multicolumn{1}{l|}{Ours}             & 31.58                    & \multicolumn{1}{c|}{0.785}   & 1.059                  & \textbf{3.81} \\ \hline
\end{tabular}
}
    \caption{Experiment results for Zh$\rightarrow$En MT attack. We list \textit{BLEU} for perturbed test sets generated by each adversarial example generation method, which is expect to deteriorate. An ideal adversarial example should achieve high \textit{RD} with respect to high \textit{HE}.
    }
    \label{tab:results}
\end{table}

As it is shown in Table \ref{tab:results} and \ref{tab:en_results}, 
our model stably generate adversarial examples without significant change in semantics \textit{by the same training setting} among different models and language pairs, achieving stably high \textit{HE}~($>$3.7) without any handcrafted semantic constraints, while search methods (GS) must tune for proper ratio of modification, which can hardly strike a balance between semantic constraints and degradation.
Unlike search paradigm relying on reference and victim gradients, our paradigm is model-agnostic yet still achieving comparable $RD$ with relatively high $HE$.

\begin{table}[]
\centering
\resizebox{0.45\textwidth}{!}{
\begin{tabular}{lcccc}
\hline
                                     & \multicolumn{4}{c}{En-De newstest13-16}                                                                   \\ \cline{2-5} 
                                     & \multicolumn{1}{l}{BLEU} & \multicolumn{1}{l|}{chrBLEU} & \multicolumn{1}{l}{RD$\uparrow$} & \multicolumn{1}{l}{HE$\uparrow$} \\ \hline
\multicolumn{1}{l|}{RNN-search-BPE}       & 25.35                    & \multicolumn{1}{c|}{-}        &        -                &  -                      \\
\multicolumn{1}{l|}{RSNI (0.2)$^*$}     & 16.70                    & \multicolumn{1}{c|}{0.949}   & 6.691$^*$                  & 2.32$^*$                   \\
\multicolumn{1}{l|}{RSNI (0.4)$^*$}     & 10.05                    & \multicolumn{1}{c|}{0.897}   & 5.860$^*$                  & 1.58$^*$                   \\
\multicolumn{1}{l|}{GS (0.2)}     & 19.42	                    & \multicolumn{1}{c|}{0.881}	 & 1.966	              & 3.81                  \\
\multicolumn{1}{l|}{GS (0.4)}     & 9.27                     & \multicolumn{1}{c|}{0.680}   & 1.982                  & 3.01                   \\
\multicolumn{1}{l|}{Ours}            & 21.27                    & \multicolumn{1}{c|}{0.921}   & \textbf{2.037}         & \textbf{3.95}          \\ \hline
\multicolumn{1}{l|}{Transformer-BPE} & 29.05                    & \multicolumn{1}{c|}{-}        &      -                  &  -                      \\
\multicolumn{1}{l|}{RSNI (0.2)$^*$}     & 18.775                   & \multicolumn{1}{c|}{0.949}   & 6.935$^*$                  & 2.39$^*$                   \\
\multicolumn{1}{l|}{RSNI (0.4)$^*$}     & 11.125                   & \multicolumn{1}{c|}{0.897}   & 5.991$^*$                  & 1.58$^*$                   \\
\multicolumn{1}{l|}{GS (0.2)}     & 18.29                       &	\multicolumn{1}{c|}{0.861}	 & 2.665	              & 3.69                  \\
\multicolumn{1}{l|}{GS (0.4)}     & 10.03                    & \multicolumn{1}{c|}{0.751}   & 2.629                  & 3.33                   \\
\multicolumn{1}{l|}{Ours}            & 19.29                    & \multicolumn{1}{c|}{0.875}   & \textbf{2.688}         & \textbf{3.79}          \\ \hline
                                     & \multicolumn{4}{c}{En-Fr newstest13-14 + newsdiscuss15}                                                   \\ \hline
\multicolumn{1}{l|}{RNN-search-BPE}       & 32.6                     & \multicolumn{1}{c|}{-}        &    -                    &   -                     \\
\multicolumn{1}{l|}{RSNI (0.2)$^*$}     & 21.93                    & \multicolumn{1}{c|}{0.947}   & 6.175$^*$                  & 2.23$^*$                  \\
\multicolumn{1}{l|}{RSNI (0.4)$^*$}     & 14.3                     & \multicolumn{1}{c|}{0.894}   & 5.271$^*$                  & 1.56$^*$                  \\
\multicolumn{1}{l|}{GS (0.2)}     & 22.7	                    & \multicolumn{1}{c|}{0.833}	  &1.818	              &3.80                  \\
\multicolumn{1}{l|}{GS (0.4)}     & 15.2                     & \multicolumn{1}{c|}{0.708}   & 1.828                  & 3.25                  \\
\multicolumn{1}{l|}{Ours}            & 22.3                     & \multicolumn{1}{c|}{0.843}   & \textbf{2.009}                  & \textbf{3.87}                  \\ \hline
\multicolumn{1}{l|}{Transformer-BPE} & 34.7                     & \multicolumn{1}{c|}{-}        &            -          &         -               \\
\multicolumn{1}{l|}{RSNI (0.2)$^*$}     & 24.0                     & \multicolumn{1}{c|}{0.947}   & 5.774$^*$                  & 2.34$^*$                  \\
\multicolumn{1}{l|}{RSNI (0.4)$^*$}     & 15.8                     & \multicolumn{1}{c|}{0.894}   & 5.114$^*$                  & 1.67$^*$                  \\
\multicolumn{1}{l|}{GS (0.2)}     & 23.01                    &   	\multicolumn{1}{c|}{0.830}	 & 1.982	              & 3.74                 \\
\multicolumn{1}{l|}{GS (0.4)}     & 19.6                     & \multicolumn{1}{c|}{0.788}   & \textbf{2.053}         & 3.68                  \\
\multicolumn{1}{l|}{Ours}            & 21.33                    & \multicolumn{1}{c|}{0.798}   & 1.907                  & \textbf{3.78}         \\ \hline
\end{tabular}
}
\caption{Experiment results for En$\rightarrow$De and En$\rightarrow$Fr MT attack.}
\label{tab:en_results}
\end{table}


\subsection{Case Study}

As it is shown in Table \ref{tab:case_study}, our method is less likely to perturb some easily-modified \textit{semantics} (e.g. numbers are edited to other ``forms'', but not different numbers), while search tends to generate semantically different tokens to achieve degradation. 
Thus our agent can lead to more insightful and plausible analyses for neural machine translation than search by gradient.

\begin{table*}[t]
    \centering
    \resizebox{\textwidth}{!}{
        \begin{tabular}{c|l}
        \hline
        \multicolumn{2}{c}{a} \\
        \hline
        \textbf{origin in} & \begin{CJK}{UTF8}{gkai}全国 4000 万选民将在16名候选人中选举法兰西第五共和国第七任总统。\end{CJK} \\
        \textbf{origin out} & 40 million voters throughout the country will elect the seventh president of the fifth republic of france among the 16 candidates \\
        \hline 
        \multirow{4}{*}{\textbf{references}} & 40 million voters in the nation will elect the 7th president for the french fifth republic from 16 candidates. \\
        & there are 40 million voters and they have to pick the fifth republic  france's seventh president amongst the sixteen candidates.\\
        & forty million voters across the country are expected to choose the 7th president of the 5th republic of france from among 16 candidates. \\
        & 40 million voters around france are to elect the 7th president of the 5 republic of france from 16 candidates .\\
        \hline
        \textbf{GS (0.4) in} & \begin{CJK}{UTF8}{gkai}全国 \textit{\textbf{性}} 4000 万\textit{\textbf{市}}民将在 \textit{\textbf{6}} 名候选人中选举法兰西第五\textit{\textbf{国家}} 第七任 \textit{\textbf{外交部长}}。 \end{CJK}\\
        \textbf{GS (0.4) out} & of the 6 candidates, 40 million people will elect the seventh foreign minister of the five countries. \\
        \hline 
        \textbf{ours in} & \begin{CJK}{UTF8}{gkai}全国\textit{\textbf{性}}4000万选民将在16\textbf{\textit{位}}候选人中选举法兰西第\textit{\textbf{5}}共和国第\textit{\textbf{7}}任总统\end{CJK} \\
        \textbf{ours out} & among the 16 candidates , 40 million voters will elect five presidents of France and seven presidents of the republic of France. \\
        \hline
        \hline
        \multicolumn{2}{c}{b} \\
        \hline
        \textbf{origin in} & \begin{CJK}{UTF8}{gkai}干案者目前被也门当局扣留。\end{CJK}\\
        \textbf{origin out} & the persons involved in the case are currently detained by the yemeni authorities.\\
        \hline
        \multirow{4}{*}{\textbf{references}}& the perpetrator is currently in the custody of the yemeni authorities. \\
        & yemeni authority apprehended the suspect. \\
        & the suspect is now in custody of yemeni authorities . \\
        & the ones involed in this case were also detained by the authority. \\
        \hline
        \textbf{GS (0.4) in} & \begin{CJK}{UTF8}{gkai}干案者目前\textit{\textbf{为}}也门\textit{\textbf{现}}局留。\end{CJK}\\
        \textbf{GS (0.4) out} & the person involved in the case is now detained by the authorities!\\
        \hline
        \textbf{ours in} & \begin{CJK}{UTF8}{gkai}干案\textit{\textbf{方}} 目前被也门当局扣留。\end{CJK} \\
        \textbf{ours out} & the victim is currently detained by the yemeni authorities.\\
        \hline
        \end{tabular}
    }
    \caption{(a) an example of perturbed number and quantifier severely damaging outputs in Zh$\rightarrow$En translation, where we highlight the changes. ``\begin{CJK}{UTF8}{gkai}五\end{CJK}''  is the character for $5$ and ``\begin{CJK}{UTF8}{gkai}七\end{CJK}'' for $7$, ``\begin{CJK}{UTF8}{gkai}名\end{CJK}'' and ``\begin{CJK}{UTF8}{gkai}位\end{CJK}'' are both commonly used quantifiers for people. However, search-based attack achieves degradation by some significant changes of semantics, where number ``16'' is changed to ``6'', and ``\begin{CJK}{UTF8}{gkai}外交部长\end{CJK}'' means ``foreign minister''. (b) an example of changed suffix which breaks the result. ``\begin{CJK}{UTF8}{gkai}方\end{CJK}'' and ``\begin{CJK}{UTF8}{gkai}者\end{CJK}'' are common suffixes (K) sharing same meaning used for people. Our model spots that victim model's fragility upon such perturb, while search does not.
    }
    \label{tab:case_study}
\end{table*}

\section{Analysis} \label{analysis}



\subsection{Efficiency}

\begin{figure}[!h]
    \centering
    \includegraphics[width=0.45\textwidth]{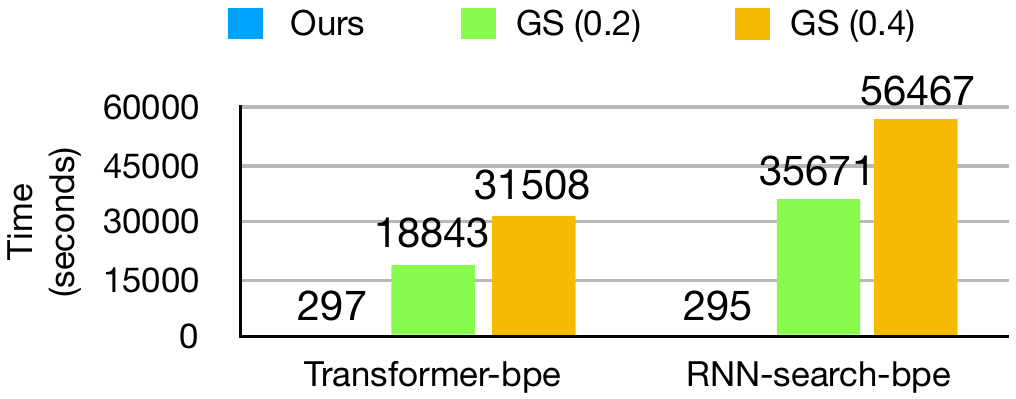}
    \caption{Time consumption of different methods: we limit memory usage to 2.5G on single Nvidia 1080, and generate adversarial examples for the same $800$ inputs in Zh$\rightarrow$En MT with different methods, our method significantly outclasses the state-of-the-art search paradigm (GS).}
    \label{fig:time_cost}
\end{figure}

As it is shown in Figure \ref{fig:time_cost}, given the same amount of memory cost, our method is significantly more efficient compared to the search paradigm. Gradient computation concerning every modified source sequence can cost considerably in time or space for a state-of-the-art system, which could be even worse for systems with recurrent units. 
When it comes to mass production of adversarial examples for a victim translation system, our method can also generate by given only monolingual inputs. In contrast, search methods must be provided the same amount of well-informed targets.

\begin{figure*}[t]
    \centering
    \includegraphics[width=\textwidth]{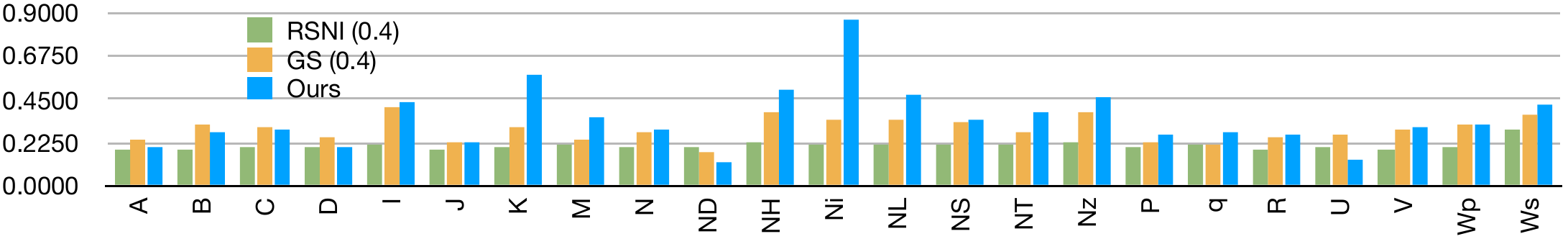}
    \caption{Attack preferences of different paradigms targeting Zh$\rightarrow$En Transformer-BPE model. All share a similar modification rate. Our agent shows a significant preference for some POS (e.g., Ni, Nh, Nz, I), which are commonly regarded as hard-to-translate phrases among industrial implementations, while some (e.g., K) are less noticed.
Preference among different choices.}
    \label{fig:attack_pattern}
\end{figure*}

\subsection{Attack Patterns}
NMT systems may have different robustness over different parts of the inputs, thus some researchers implement input preprocessing targeting certain empirically weak parts, e.g., named entities\citep{li-etal-2018-named}.
Since the agent's policy is to attack without handcrafted error features, we can further investigate vulnerability by its attack preferences of different parts of speech.
We choose Chinese, for example, and adopt LTP POS tagger\footnote{https://github.com/HIT-SCIR/ltp} to label NIST test sets, then check the modification rate for each POS. To ensure the reliability of our analysis, we run three rounds of experiments on both baselines and our agent with \textit{similar modification rate} targeting state-of-the-art Transformer with BPE, and collect overall results. We also present random synthetic noise injection \citep{karpukhin2019training} (RSNI), which is not intended for any preference as an additional baseline. 

As it is shown in Figure \ref{fig:attack_pattern}, our reinforced paradigm shows distinct preference upon certain POS tags, indicating pitfalls of a victim translation system. 
At the same time, RSNI distributed almost evenly upon different POS tags. 
Though the search paradigm (GS) does expose some types of pitfall, our method can further expose those omitted by the search.
Note that unlike existing work relying on feature engineering to indicate errors, we have no such features implemented for an agent. However, our agent can still spot error patterns by favoring some of the POS, such as Ni (organization name), Nh (person name), Nl (location name), M (numbers), which are commonly accepted as hard-to-translate parts. Moreover, the agent also tends to favor K (suffix) more, which is less noticed. 


\subsection{Attack Generalization}
We additionally test agents by attacking different model architecture from the one that it's trained. 
As it is shown in Table \ref{tab:dis_attack}, we perturb the inputs by agents trained to attack a different architecture, then test for degradation.
The results show that our agent trained by targeting Transformer architecture can still achieve degradation on RNN-search, and vice-versa. 

\begin{table}[]
    \centering
    {
    \resizebox{0.45\textwidth}{!}{
        \begin{tabular}{lll}
\hline
                                                  & \multicolumn{1}{l|}{Attack by}     & BLEU$(\Delta)$             \\ \cline{2-3} 
                                                  & \multicolumn{2}{c}{Zh-En MT02-06}                     \\ \hline
\multicolumn{1}{l|}{\multirow{3}{*}{RNN-search-BPE}}  & \multicolumn{1}{c|}{-}             & 40.90            \\
\multicolumn{1}{l|}{}                             & \multicolumn{1}{c|}{agent-RNN}     & 31.58(-9.32)     \\
\multicolumn{1}{l|}{}                             & \multicolumn{1}{c|}{agent-TF}      & 32.14(-8.76)     \\ \hline
\multicolumn{1}{l|}{\multirow{3}{*}{Transformer-BPE}} & \multicolumn{1}{c|}{-}             & 44.06            \\
\multicolumn{1}{l|}{}                             & \multicolumn{1}{c|}{agent-TF}      & 35.48(-8.58)     \\
\multicolumn{1}{l|}{}                             & \multicolumn{1}{c|}{agent-RNN}     & 33.14(-10.92)    \\ \hline
                                                  & \multicolumn{2}{c}{En-De Newstest13-16}               \\ \hline
\multicolumn{1}{l|}{\multirow{3}{*}{RNN-search-BPE}}  & \multicolumn{1}{c|}{-}             & 25.35            \\
\multicolumn{1}{l|}{}                             & \multicolumn{1}{c|}{agent-RNN}     & 21.27(-4.08)     \\
\multicolumn{1}{l|}{}                             & \multicolumn{1}{c|}{agent-TF}      & 17.18(-8.18)     \\ \hline
\multicolumn{1}{l|}{\multirow{3}{*}{Transformer-BPE}} & \multicolumn{1}{c|}{-}             & 29.05            \\
\multicolumn{1}{l|}{}                             & \multicolumn{1}{c|}{agent-TF}      & 19.29(-9.76)     \\
\multicolumn{1}{l|}{}                             & \multicolumn{1}{c|}{agent-RNN}     & 24.2(-4.85)      \\ \hline
                                                  & \multicolumn{2}{c}{En-Fr Newstest13-14+newsdiscuss15} \\ \hline
\multicolumn{1}{l|}{\multirow{3}{*}{RNN-search-BPE}}  & \multicolumn{1}{c|}{-}             & 32.60            \\
\multicolumn{1}{l|}{}                             & \multicolumn{1}{c|}{agent-RNN}     & 22.3(-10.30)     \\
\multicolumn{1}{l|}{}                             & \multicolumn{1}{c|}{agent-TF}      & 19.83(-14.87)    \\ \hline
\multicolumn{1}{l|}{\multirow{3}{*}{Transformer-BPE}} & \multicolumn{1}{c|}{-}             & 34.70            \\
\multicolumn{1}{l|}{}                             & \multicolumn{1}{c|}{agent-TF}      & 21.33(-13.37)    \\
\multicolumn{1}{l|}{}                             & \multicolumn{1}{c|}{agent-RNN}     & 22.35(-10.25)    \\ \hline
\end{tabular}}
    \caption{Attacks targeting different architecture from the trained one. We note agent with the architecture that is trained with(e.g., agent-RNN stands for agent trained by targeting RNN-search).}
    \label{tab:dis_attack}
    }
\end{table}

\subsection{Tuning with Adversarial Examples}

Since the agent generates meaning-preserving adversarial examples efficiently, we can directly tune the original model with those samples. 
We choose Zh$\rightarrow$En Transformer-BPE, for example, and generate the same amount of adversarial examples given original training sources(1.3M pairs), then paired with initial targets.
We mix the augmented pairs with original pairs for a \textit{direct} tuning. 
We test the tuned model on original test data and noisy test data generated by the attacking agent.
We additionally test on IWSLT11-17 Zh$\rightarrow$En test data, which is not used for training or tuning, to verified robustness improvement. 
As Table \ref{tab:tuning} shows, our agent can achieve substantial improvement(+4.83) on noisy tests with only minor loss on clean tests(-0.46).
The improvement on the IWSLT test also indicates the adversarial tuning contributes to not only defending the agent's attack, but also overall robustness.

\begin{table}[]
    \centering
    \resizebox{0.45\textwidth}{!}{
    \begin{tabular}{ll|l|l}
\hline
                                     & Clean test & Noisy test & IWSLT11-17 \\ \hline
\multicolumn{1}{l|}{Transformer-BPE} & 44.06     & 35.48            & 11.27      \\ \hline
\multicolumn{1}{l|}{Tuned}           & 43.60(-0.46)     & 40.31(+4.83)            & 11.73(+0.46)      \\ \hline
\end{tabular}}
    \caption{Tuning Zh$\rightarrow$En Transformer-BPE model with adversarial examples. We generate adversarial examples for every training sources for tuning, achieving overall improvements for noisy tests.}
    \label{tab:tuning}
\end{table}

\begin{table}[!t]
    \centering
    \resizebox{0.47\textwidth}{!}{
        \begin{tabular}{c|l}
        \hline
        \textbf{in} & \begin{CJK}{UTF8}{gkai}钱其琛同突尼斯外长会谈。\end{CJK} \\
        \hline
        \textbf{perturbed in} & \begin{CJK}{UTF8}{gkai}钱其琛同突尼斯外长会谈\textbf{\textit{-}}\end{CJK} \\
        \hline
        \textbf{out} & Chinese, Tunisian minsters hold talks. \\
        \hline
        \textbf{perturbed out} & qian qichen holds talks with Tunisian foreign minister. \\
        \hline
        \hline
        \textbf{in} & \begin{CJK}{UTF8}{gkai}中巴及城巴车辆在南区通宵停泊\end{CJK} \\
        \hline
        \textbf{perturbed in} &\begin{CJK}{UTF8}{gkai}中巴及城巴车辆在南区通宵停\textbf{\textit{车}}\end{CJK} \\
        \hline
        \textbf{out} & overnight parking of cmb and city bus \\
        \hline
        \textbf{perturbed out} & overnight parking of cmb and city bus in southern district \\
        \hline
        \end{tabular}
    }
    \caption{Example of minor perturbed samples that improves machine translation for Zh$\rightarrow$En Transformer-BPE model. The ``\begin{CJK}{UTF8}{gkai}。\end{CJK}'' in first sample is modified to ``-'', then model yields the omitted ``\begin{CJK}{UTF8}{gkai}钱其琛\end{CJK} (qian qi chen)''. The ``\begin{CJK}{UTF8}{gkai}停泊\end{CJK}'' in second sample is modified to ``\begin{CJK}{UTF8}{gkai}停车\end{CJK}'', where they both mean ``parking'', then comes the omitted ``in southern district'' for ``\begin{CJK}{UTF8}{gkai}在南区\end{CJK}''. }
    \label{tab:reinforced_examples}
\end{table}

\subsection{Reinforced Examples for Machine translation}
We additionally switched the episodic rewards in the environment, then ignored all modifications that induce UNK tokens to train an agent, hoping to generate minor perturbed samples that can improve the translation metric. 
Though we failed to achieve overall improvements, we do succeed for quite a portion of samples, as shown in Table \ref{tab:reinforced_examples}. 
Similar to adversarial examples, we call them \textit{reinforced examples}. 
Such improvement is different from adversarial training that tunes model for defense or strict text correction before the test phase. 
Reinforced examples are still noisy and can be directly applied for a test without any model updates to achieve improvements, which to our best knowledge is less investigated by researchers.
Since we discovered that not all perturbed inputs are harmful, such an issue can be a good hint and alternative for better adversarial defense in NLP and should be further considered.

\section{Related Work}
\newcite{cheng2018seq2sick} and \newcite{cheng2018towards} applied continuous perturbation learning on token's embedding and then manage a lexical representation out of a perturbed embedding. \newcite{zhao2017generating} learned such perturbation on the encoded representation of a sequence, and then decode it back as an adversarial example.
These methods are applicable for simple NLP classification tasks, while failing machine translation which requires higher semantic constraints.
\newcite{zhao2017generating} further attempted to constrain semantic in such paradigm by introducing multi-task modeling with accessory annotation, which further limits applicability. 


On the other hand, \newcite{ebrahimi2018adversarial}, \newcite{Chaturvedi2019ExploringTR} and \newcite{cheng2019robust} regarded it as a search problem by maximizing surrogate gradient losses.
Due to the formidable gradient computation, such methods are less viable to more complex neural architectures. 
\newcite{cheng2019robust} introduced a learned language model to constrain generation. However, a learned language model is not apt for common typos or UNK. 
Another pitfall of this paradigm is that surrogate losses defined by a fixed tokenization for non-character level systems, risks being invalidated once the attack changes tokenization. Therefore, \newcite{ebrahimi2018adversarial} simply focused on char-level systems, while \newcite{michel2019evaluation} specially noted to exclude scenarios where attack changes tokenization in their paradigm.

Other works turn to more sophisticated generation paradigms, e.g., \newcite{Vidnerov2016Evolutionary} adopts a genetic algorithm for an evolutionary generation targeting simple machine learning models. 
\newcite{Zang2019Textual} consider adversarial generation as a word substitution-based combinatorial optimization problem tackled by particle swarm algorithm. 
Our paradigm shares some common ideology with \newcite{miao2019cgmh} and \newcite{ijcai2018-543}, which iteratively edit inputs constrained by generative adversarial learning.

\section{Conclusion}
We propose a new paradigm to generate adversarial examples for neural machine translation, which is capable of exposing translation pitfalls without handcrafted error features. 
Experiments show that our method achieves stable degradation with meaning preserving adversarial examples over different victim models. 

It is noticeable that our method can generate adversarial examples efficiently from monolingual data. 
As a result, the mass production of adversarial examples for the victim model's analysis and further improvement of robustness become convenient. 
Furthermore, we notice some exceptional cases which we call as ``reinforced samples'', which we leave as the future work.

\section*{Acknowledgement}
We would like to thank the anonymous reviewers for their insightful comments. Shujian Huang is the corresponding author. This work is supported by National Science Foundation of China (No. 61672277, 61772261),  the Jiangsu Provincial Research Foundation for Basic Research (No. BK20170074).

\bibliographystyle{acl_natbib}
\bibliography{reference}

\begin{thebibliography}{31}
\expandafter\ifx\csname natexlab\endcsname\relax\def\natexlab#1{#1}\fi

\bibitem[{Bahdanau et~al.(2016)Bahdanau, Brakel, Xu, Goyal, Lowe, Pineau,
  Courville, and Bengio}]{bahdanau2016actor}
Dzmitry Bahdanau, Philemon Brakel, Kelvin Xu, Anirudh Goyal, Ryan Lowe, Joelle
  Pineau, Aaron Courville, and Yoshua Bengio. 2016.
\newblock An actor-critic algorithm for sequence prediction.
\newblock \emph{arXiv preprint arXiv:1607.07086}.

\bibitem[{Bahdanau et~al.(2014)Bahdanau, Cho, and Bengio}]{Bahdanau2015Neural}
Dzmitry Bahdanau, Kyunghyun Cho, and Yoshua Bengio. 2014.
\newblock Neural machine translation by jointly learning to align and
  translate.
\newblock \emph{CoRR}.

\bibitem[{Belinkov and Bisk(2017)}]{belinkov2017synthetic}
Yonatan Belinkov and Yonatan Bisk. 2017.
\newblock Synthetic and natural noise both break neural machine translation.
\newblock \emph{arXiv preprint arXiv:1711.02173}.

\bibitem[{Chaturvedi et~al.(2019)Chaturvedi, Abijith, and
  Garain}]{Chaturvedi2019ExploringTR}
Akshay Chaturvedi, KP~Abijith, and Utpal Garain. 2019.
\newblock Exploring the robustness of nmt systems to nonsensical inputs.
\newblock \emph{arXiv: Learning}.

\bibitem[{Cheng et~al.(2018{\natexlab{a}})Cheng, Yi, Zhang, Chen, and
  Hsieh}]{cheng2018seq2sick}
Minhao Cheng, Jinfeng Yi, Huan Zhang, Pin-Yu Chen, and Cho-Jui Hsieh.
  2018{\natexlab{a}}.
\newblock Seq2sick: Evaluating the robustness of sequence-to-sequence models
  with adversarial examples.
\newblock \emph{arXiv preprint arXiv:1803.01128}.

\bibitem[{Cheng et~al.(2019)Cheng, Jiang, and Macherey}]{cheng2019robust}
Yong Cheng, Lu~Jiang, and Wolfgang Macherey. 2019.
\newblock Robust neural machine translation with doubly adversarial inputs.
\newblock \emph{arXiv preprint arXiv:1906.02443}.

\bibitem[{Cheng et~al.(2018{\natexlab{b}})Cheng, Tu, Meng, Zhai, and
  Liu}]{cheng2018towards}
Yong Cheng, Zhaopeng Tu, Fandong Meng, Junjie Zhai, and Yang Liu.
  2018{\natexlab{b}}.
\newblock Towards robust neural machine translation.
\newblock \emph{arXiv preprint arXiv:1805.06130}.

\bibitem[{Ebrahimi et~al.(2018)Ebrahimi, Lowd, and
  Dou}]{ebrahimi2018adversarial}
Javid Ebrahimi, Daniel Lowd, and Dejing Dou. 2018.
\newblock On adversarial examples for character-level neural machine
  translation.
\newblock \emph{arXiv preprint arXiv:1806.09030}.

\bibitem[{Goodfellow et~al.(2014)Goodfellow, Shlens, and
  Szegedy}]{goodfellow2014explaining}
Ian~J Goodfellow, Jonathon Shlens, and Christian Szegedy. 2014.
\newblock Explaining and harnessing adversarial examples.
\newblock \emph{arXiv preprint arXiv:1412.6572}.

\bibitem[{Karpukhin et~al.(2019)Karpukhin, Levy, Eisenstein, and
  Ghazvininejad}]{karpukhin2019training}
Vladimir Karpukhin, Omer Levy, Jacob Eisenstein, and Marjan Ghazvininejad.
  2019.
\newblock Training on synthetic noise improves robustness to natural noise in
  machine translation.
\newblock \emph{arXiv preprint arXiv:1902.01509}.

\bibitem[{Konda and Tsitsiklis(2000)}]{konda2000actor}
Vijay~R Konda and John~N Tsitsiklis. 2000.
\newblock Actor-critic algorithms.
\newblock In \emph{Advances in neural information processing systems}, pages
  1008--1014.

\bibitem[{Lee et~al.(2018)Lee, Firat, Agarwal, Fannjiang, and
  Sussillo}]{lee2018hallucinations}
Katherine Lee, Orhan Firat, Ashish Agarwal, Clara Fannjiang, and David
  Sussillo. 2018.
\newblock Hallucinations in neural machine translation.
\newblock \emph{NIPS 2018 Workshop IRASL}.

\bibitem[{Li et~al.(2018)Li, Wang, Aw, Chng, and Li}]{li-etal-2018-named}
Zhongwei Li, Xuancong Wang, Ai~Ti Aw, Eng~Siong Chng, and Haizhou Li. 2018.
\newblock \href {https://doi.org/10.18653/v1/W18-2407} {Named-entity tagging
  and domain adaptation for better customized translation}.
\newblock In \emph{Proceedings of the Seventh Named Entities Workshop}, pages
  41--46, Melbourne, Australia. Association for Computational Linguistics.

\bibitem[{Luong et~al.(2015)Luong, Pham, and Manning}]{Luong2015Effective}
Minh{-}Thang Luong, Hieu Pham, and Christopher~D. Manning. 2015.
\newblock Effective approaches to attention-based neural machine translation.
\newblock In \emph{EMNLP}.

\bibitem[{Miao et~al.(2019)Miao, Zhou, Mou, Yan, and Li}]{miao2019cgmh}
Ning Miao, Hao Zhou, Lili Mou, Rui Yan, and Lei Li. 2019.
\newblock Cgmh: Constrained sentence generation by metropolis-hastings
  sampling.
\newblock In \emph{Proceedings of the AAAI Conference on Artificial
  Intelligence}, volume~33, pages 6834--6842.

\bibitem[{Michel et~al.(2019)Michel, Li, Neubig, and
  Pino}]{michel2019evaluation}
Paul Michel, Xian Li, Graham Neubig, and Juan~Miguel Pino. 2019.
\newblock On evaluation of adversarial perturbations for sequence-to-sequence
  models.
\newblock \emph{arXiv preprint arXiv:1903.06620}.

\bibitem[{Mnih et~al.(2015)Mnih, Kavukcuoglu, Silver, Rusu, Veness, Bellemare,
  Graves, Riedmiller, Fidjeland, Ostrovski et~al.}]{mnih2015human}
Volodymyr Mnih, Koray Kavukcuoglu, David Silver, Andrei~A Rusu, Joel Veness,
  Marc~G Bellemare, Alex Graves, Martin Riedmiller, Andreas~K Fidjeland, Georg
  Ostrovski, et~al. 2015.
\newblock Human-level control through deep reinforcement learning.
\newblock \emph{Nature}, 518(7540):529.

\bibitem[{Post(2018)}]{post2018call}
Matt Post. 2018.
\newblock A call for clarity in reporting bleu scores.
\newblock \emph{arXiv preprint arXiv:1804.08771}.

\bibitem[{Sano et~al.(2019)Sano, Suzuki, and Kiyono}]{sano2019effective}
Motoki Sano, Jun Suzuki, and Shun Kiyono. 2019.
\newblock Effective adversarial regularization for neural machine translation.
\newblock In \emph{Proceedings of the 57th Annual Meeting of the Association
  for Computational Linguistics}, pages 204--210.

\bibitem[{Schulman et~al.(2015)Schulman, Moritz, Levine, Jordan, and
  Abbeel}]{schulman2015high}
John Schulman, Philipp Moritz, Sergey Levine, Michael Jordan, and Pieter
  Abbeel. 2015.
\newblock High-dimensional continuous control using generalized advantage
  estimation.
\newblock \emph{arXiv preprint arXiv:1506.02438}.

\bibitem[{Sennrich et~al.(2015)Sennrich, Haddow, and
  Birch}]{sennrich2015neural}
Rico Sennrich, Barry Haddow, and Alexandra Birch. 2015.
\newblock Neural machine translation of rare words with subword units.
\newblock \emph{arXiv}.

\bibitem[{Shazeer and Stern(2018)}]{shazeer2018adafactor}
Noam Shazeer and Mitchell Stern. 2018.
\newblock Adafactor: Adaptive learning rates with sublinear memory cost.
\newblock \emph{arXiv preprint arXiv:1804.04235}.

\bibitem[{Sutton and Barto(2018)}]{sutton2018reinforcement}
Richard~S Sutton and Andrew~G Barto. 2018.
\newblock \emph{Reinforcement learning: An introduction}.
\newblock MIT press.

\bibitem[{Vaswani et~al.(2017)Vaswani, Shazeer, Parmar, Uszkoreit, Jones,
  Gomez, Kaiser, and Polosukhin}]{vaswani2017attention}
Ashish Vaswani, Noam Shazeer, Niki Parmar, Jakob Uszkoreit, Llion Jones,
  Aidan~N Gomez, {\L}ukasz Kaiser, and Illia Polosukhin. 2017.
\newblock Attention is all you need.
\newblock In \emph{NIPS}.

\bibitem[{Vidnerová and Neruda(2016)}]{Vidnerov2016Evolutionary}
Petra Vidnerová and Roman Neruda. 2016.
\newblock Evolutionary generation of adversarial examples for deep and shallow
  machine learning models.
\newblock In \emph{Multidisciplinary International Social Networks Conference}.

\bibitem[{Wu et~al.(2018)Wu, Tian, Qin, Lai, and Liu}]{wu2018study}
Lijun Wu, Fei Tian, Tao Qin, Jianhuang Lai, and Tie-Yan Liu. 2018.
\newblock A study of reinforcement learning for neural machine translation.
\newblock \emph{arXiv preprint arXiv:1808.08866}.

\bibitem[{Xiao et~al.(2018)Xiao, Li, yan Zhu, He, Liu, and
  Song}]{ijcai2018-543}
Chaowei Xiao, Bo~Li, Jun yan Zhu, Warren He, Mingyan Liu, and Dawn Song. 2018.
\newblock \href {https://doi.org/10.24963/ijcai.2018/543} {Generating
  adversarial examples with adversarial networks}.
\newblock In \emph{Proceedings of the Twenty-Seventh International Joint
  Conference on Artificial Intelligence, {IJCAI-18}}, pages 3905--3911.
  International Joint Conferences on Artificial Intelligence Organization.

\bibitem[{Zang et~al.(2019)Zang, Yang, Qi, Liu, Zhang, Liu, and
  Sun}]{Zang2019Textual}
Yuan Zang, Chenghao Yang, Fanchao Qi, Zhiyuan Liu, Meng Zhang, Qun Liu, and
  Maosong Sun. 2019.
\newblock Textual adversarial attack as combinatorial optimization.
\newblock \emph{arXiv:1910.12196v2}.

\bibitem[{Zhao et~al.(2018)Zhao, Zhang, He, Zong, and Wu}]{zhao2018addressing}
Yang Zhao, Jiajun Zhang, Zhongjun He, Chengqing Zong, and Hua Wu. 2018.
\newblock Addressing troublesome words in neural machine translation.
\newblock In \emph{Proceedings of the 2018 Conference on Empirical Methods in
  Natural Language Processing}, pages 391--400.

\bibitem[{Zhao et~al.(2017)Zhao, Dua, and Singh}]{zhao2017generating}
Zhengli Zhao, Dheeru Dua, and Sameer Singh. 2017.
\newblock Generating natural adversarial examples.
\newblock \emph{arXiv preprint arXiv:1710.11342}.

\bibitem[{Ziebart(2010)}]{ziebart2010modeling}
Brian~D Ziebart. 2010.
\newblock \emph{Modeling purposeful adaptive behavior with the principle of
  maximum causal entropy}.
\newblock Ph.D. thesis, figshare.

\end{thebibliography}

\appendix
\section{Training Details for Agent}

We adopt commonly accepted translation metric BLEU as $score$ in Eq.\ref{eq:r_degrade}.
We use 50 sequence pairs per batch both in environment initialization and training of discriminator and agent.
It is essential to train on batches of sequences to stabilize reinforced training.
Furthermore, note that $D$ can be too powerful during the early training stage compared to the agent's actor that it can quickly terminate an exploration. Therefore, we must train on batches and determine an overall terminal signal as aforementioned to ensure early exploration.
The $step_D$ and $step_A$ are set as $80$ \footnote{ Three times the average convergence episodes to train a discriminator with initial agent by the given batch size.} and $120$. $acc\_bound$ for discriminator training is set to $0.85$. The $a$ and $b$ in Eq.\ref{eq:rewards} are set to $0.5$ and $10$. 
The dimension of feedforward layers in the agent's actor-critic and discriminator are all $256$.
We initialize the embedding of both agent and discriminator by the victim's embedding.

For reinforcement learning, we adopt asynchronous learning with an additional global agent with an additional set of parameter $\theta^{\Omega}$, we set discount factor $\gamma$ to $0.99$, $\alpha$ and $\beta$ in Eq.\ref{eq:overall_loss} to $0.5$ and $0.05$ respectively. 
As for the stop criterion, we set $patience\_round$ to $15$ with convergence boundary for $acc_D$ to $0.52$.
We adopt Adafactor\citep{shazeer2018adafactor} for training, which is a memory-efficient Adam. The learning rate for agent's optimizer is initiated as $0.001$ and scheduled by rsqrt with 100 steps of warmup. The $K$ for the candidate set is 12.

Our agent takes around 30 hours to converge on a single Nvidia 1080ti. 
\label{sec:Alg_train}
\begin{algorithm}[!ht]
  \SetAlgoLined
  \KwResult{A learned global agent $\pi_{\theta^{\Omega}}$}
  \text{Assume global agent as $\pi_{\theta^{\Omega}}$ with parameter $\theta^{\Omega}$ } \\
  \text{Assume agent as $\pi_\theta$ with parameter set $\theta$} \\
  initialize: $Env$ with $D$, $\theta^{\Omega}$, $\theta$ \;
  \While{\text{not Stop Criterion}}{
    \For{$step_D$}{
     train $D$ with current agent $\pi_\theta$ \;
     if {$acc_D > acc\_bound$ }
        break\;
    }
    test current $D$'s accuracy $acc_D$ for stop criterion\;
    \For{$step_A$}{ 
      initialize $Env$ state $s_0$\; 
      synchronize $\pi_\theta$ with $\pi_{\theta^{\Omega}}$ \;
      $t=t_{start}$ \;
      \While{\text{$s_t$ survive and $t-t_{start}<t_{max}$}}
      {get $out_t^{\text{actor}}, V_t=\pi_\theta(s_t)$ \;
       compute entropy $H(out_t^{\text{actor}})$ \;
       sample $a_t$ based on $out_t^{\text{actor}}$ \;
       perform $a_t$ and receive $r_t$ and $s_{t+1}$ \;
       $t\leftarrow t+1$ \;
      }
      $R=\begin{cases}
        0   & \text{for terminal $s_t$} \\
        V(s_t)  &\text{for non-terminal $s_t$} \\
        \end{cases}$  \\
      \For{$i\in\{t-1,...,t_{start}\}$ }
      {  $R\leftarrow \gamma R + r_i $ \;
         accumulate $L_t^v(\theta)$ \;
         accumulate $L_t^{\pi}(\theta)$ \;
      }
      compute overall loss $L(\theta)$ \;
      perform asynchronous updates on $\theta^{\Omega}$ with gradient $\frac{\partial L(\theta)}{\partial \theta}$ \;
    }
   }
  \caption{Reinforced training for agent}
  \label{alg:train_overview}
\end{algorithm}
Note that higher $acc\_bound$ and lower convergence boundary for $D$ indicates higher semantic constraints, which will increase training time. 

\section{Search-based Attack}
Search-based adversarial generation is currently widely applied in various robustness machine translation system. 
We generally follow the strategy of \newcite{ebrahimi2018adversarial,michel2019evaluation}  which is applicable for both RNN-search and Transformer. More specifically, the $L_{adv}$ in Eq.\ref{eq:adversarial_loss_discrete} is derived as:

\begin{align}
\centering
    &\underset{1 \leq i \leq n, emb_i' \in vocab}{\mathrm{argmax}}|emb'-emb_i| \nabla _{emb_i}L_{adv}, \\
    &L_{adv}(X',Y) = \sum^{|y|}_{t=1} log(1-P(y_t|X',y_{<t-1}))\nonumber
\end{align}
where each $P(y_t|X)$ is calculated by Eq.\ref{eq:nmt_output} given a corresponding target.
For every source sequence, a small ratio of positions is sampled for search. Then we greedy search\footnote{\newcite{ebrahimi2018adversarial} suggest that greedy search is a good enough \textbf{approximation}.} by the corresponding loss upon those positions with given candidates.
For better comparison, we adopt the candidate set used in our model instead of naive KNN candidates. Both baseline and our model share the same UNK generation for presentation. We use homophone replacement for Chinese, and strategy by \newcite{michel2019evaluation} for English.

\end{document}